%% file: main.tex
\documentclass{article} 
\usepackage{iclr_r2fm,times}

\input{math_commands.tex}

\usepackage{hyperref}
\usepackage{url}
\usepackage[utf8]{inputenc} 
\usepackage[T1]{fontenc}    
\usepackage{hyperref}       
\usepackage{url}            
\usepackage{booktabs}       
\usepackage{amsfonts}       
\usepackage{nicefrac}       
\usepackage{microtype}      
\usepackage{xcolor}         
\usepackage{tabularx}
\usepackage{multirow}
\usepackage{colortbl}
\usepackage{graphicx}
\usepackage{subcaption}
\definecolor{lightgray}{gray}{0.8}

\newcommand{\Caner}[1]{{\color{blue}[]}}

\title{The bias of harmful label associations in vision-language models}

\author{Caner Hazirbas, Alicia Sun,  Yonathan Efroni, Mark Ibrahim \\
Meta AI
}

\iclrfinalcopy 
\begin{document}

\maketitle

\begin{abstract}

Despite the remarkable performance of foundation vision-language models, 
the shared representation space for text and vision can also encode harmful label associations
detrimental to fairness. 
While prior work has uncovered bias in vision-language models' (VLMs) classification performance across geography, work has been limited
along the important axis of harmful label associations due to a lack of rich, labeled data.
In this work, we investigate harmful label associations in the recently released Casual Conversations datasets containing more than 70,000 videos.
We study bias in the frequency of harmful label associations across self-provided labels for age, gender, apparent skin tone, and physical adornments across several leading VLMs. We find that VLMs are $4-7$x more likely to harmfully classify individuals with darker skin tones. 
We also find scaling transformer encoder model size leads to higher confidence in harmful predictions. 
Finally, we find improvements on standard vision tasks across VLMs does not address disparities in harmful label associations.

\end{abstract}

\section{Introduction}

Vision-language models (VLMs) offer a new capability for zero-shot open-world object recognition: the ability to classify any object without additional training.
Compared to standard supervised training where the set of classes is predefined, VLMs can classify any text-image pairs thanks to a shared embedding space for images and text. 
With this new capability however, comes the potential for VLMs to learn a multidude of harmful spurious correlations between text and images. 

While prior work attempted to study such biases in the context of domain shifts \citep{kalluri2023geonet}, studies specifically tackling these sensitive axes have been limited due to dataset challenges \citep{richards2023does}. 
Few, if any datasets, contain large diverse visuals across these many axes of gender, geography etc. and when they do, they often lack ground truth labels. For example, DollarStreet contains 30k samples with ground truth labels only for income and geography. Here we study these critical axes using the recently released Casual Conversations 
datasets~\cite{hazirbas2022ccv1,porgali2023casual}. 
Casual Conversations contains more than $70,000$ videos of $8,500$ individuals with self-provided age, gender, physical attributes \& adornments and geo-location and apparent skin tone. 

Our study focuses on vision-language model predictions associating a human with a harmful class label (on the standard 1k ImageNet classes)---in other words, a \textit{harmful prediction}. We study several vision-language models pre-trained on web scale data that perform well on standard classification benchmarks, including CLIP~\cite{radford2021learning} of various transformer encoder sizes and BLIP-2~\citep{li2023blip2} and other variants~\citep{li2022blip}. We measure the extent harmful predictions disproportionately affect some groups more than others across age, gender, and apparent skin tone as well as the effect of physical adornments such as glasses and hats.
We find 1) VLMs are $4-7$x more likely to harmfully classify individuals with darker skin tones relative those with lighter skin tones, 2) larger transformer models are more confident in their harmful label associations, and 3) progress on standard vision tasks does not necessarily improve disparities in harmful label associations. 

Recent work has shown that vision-language model could induce harms by amplifying societal biases  \citep{Agarwal2021EvaluatingCT, Hamidieh2023Identifying, Zhan2022Counterfactually, Hirota2023Model}, and causing performance disparities across demographic groups \citep{Hall2023Vision}, yet these analysis were only conducted on specific demographic groups. In this work, we investigate harmful label associations across a more comprehensive set of axes.



\section{Methodology}

We evaluate bias in vision-language models on geographically diverse video datasets of humans from the Casual Conversations v1 and v2 datasets. 
We compare harmful label associations of several leading vision-language models across self-identified attributes such as age, gender and apparent skin tone as well as physical adornments including face mask, beard/mustache, hair cover, etc. 

In our experiments, we use the both versions of the Casual Conversations datasets~\cite{hazirbas2022ccv1,porgali2023casual}. Both datasets provide self-identified age, gender and apparent skin tone per participant, while CCv2 is richer and more diverse in terms of participant and labels. CCv1 is composed of 3,011 participants collected across five cities in the US. CCv2 has 5,567 participants from seven different countries. We processed videos as described 
in~\ref{apx:data}.

We focus this study on foundation models such as CLIP which encode both vision and text in the same representation space. 
We compare CLIP models with ViT, transformer encoders, of varying sizes including B16, B32, and L14. We also study a more recent vision-language model, BLIP2, trained with additional captioning and image-text matching objectives. To classify an image, we encode both the image and text prompts for each class label then predict 
the class label with the highest similarity to our image representation following the same procedure in~\cite{radford2021learning}.

\begin{figure}
    \centering
    \begin{subfigure}[t]{0.45\textwidth}
        \centering
        \includegraphics[width=\textwidth]{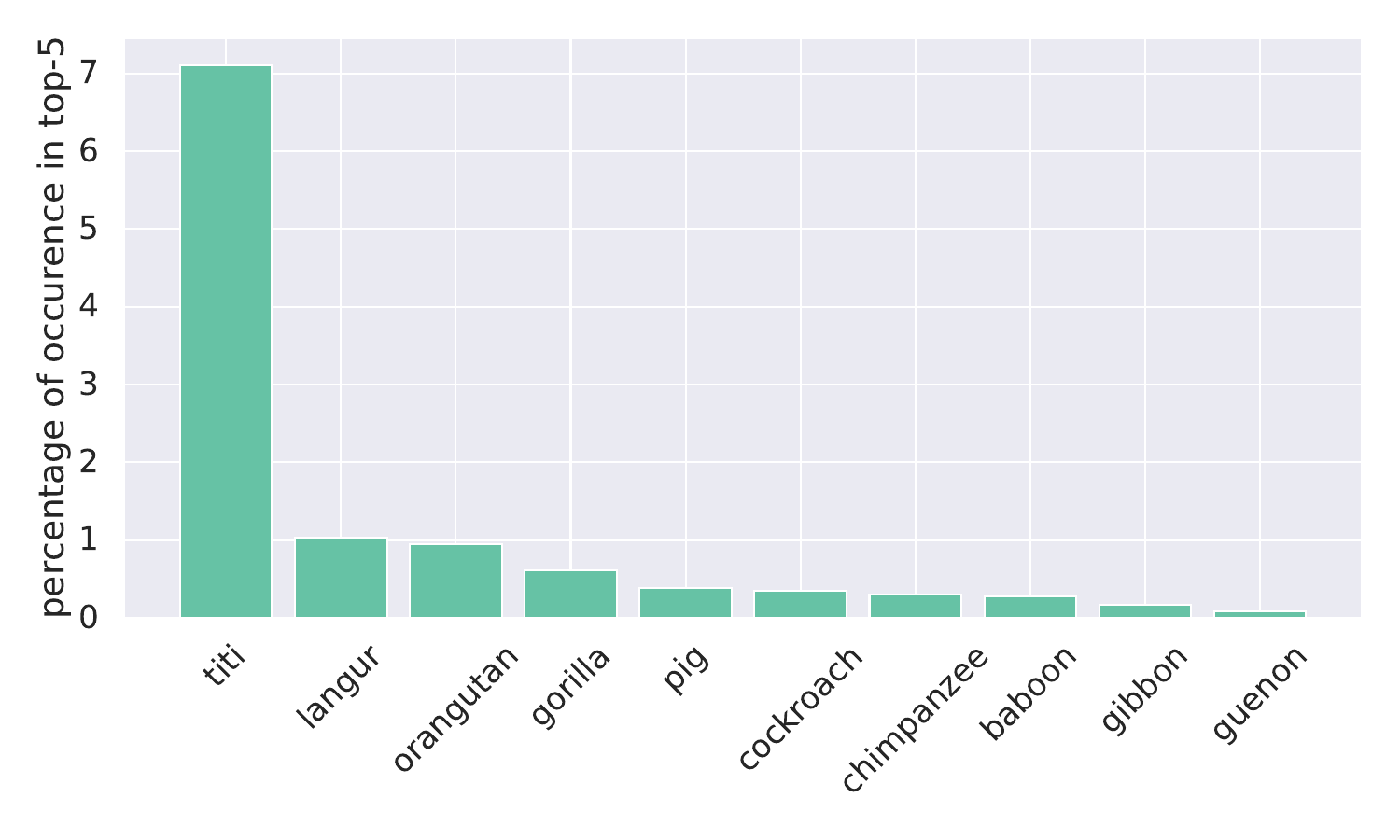}
    \end{subfigure}
    \begin{subfigure}[t]{0.44\textwidth}
    \centering
    \includegraphics[width=0.9\textwidth]{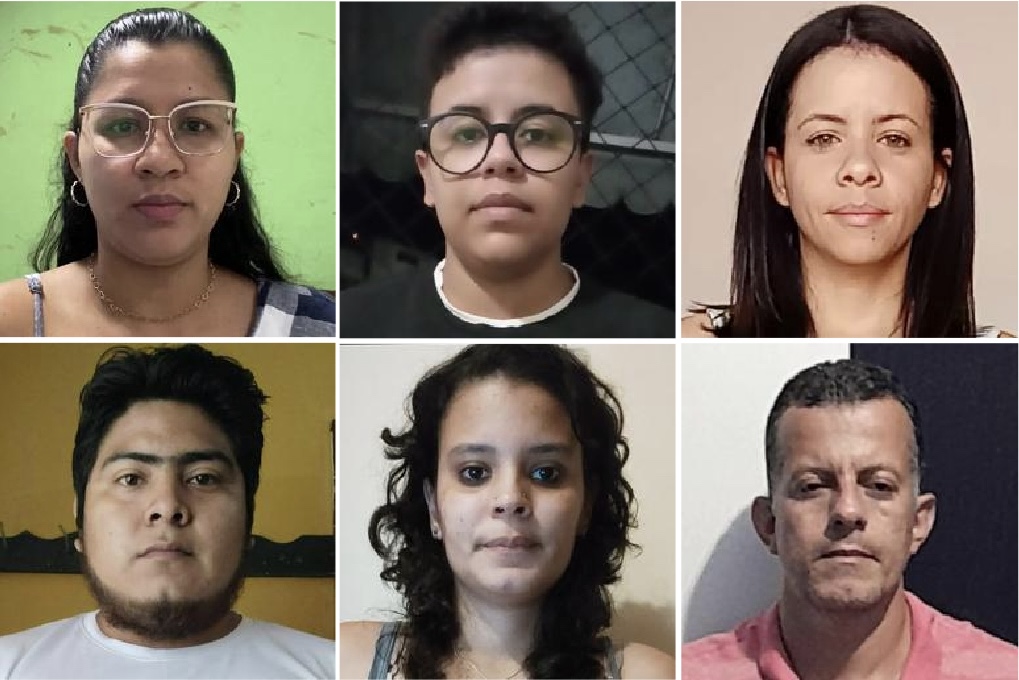}
    \end{subfigure}
    \caption{Most commonly selected harmful class labels for CLIP ViT-L14 (left) along with sample frames leading to harmful label associations from CCv2 (right).}
    \label{fig:most-common-harmful-labels}
\end{figure}

We focus our study of bias on harmful label associations: model label predictions that inappropriately classify a human as another harmful label, such as a primate. We use a standard zero-shot classification across the standard 1k ImageNet classes. 
We use the 80 text prompts~\footnote{\tiny \url{https://github.com/openai/CLIP/blob/main/notebooks/Prompt_Engineering_for_ImageNet.ipynb}} with simplified 1k ImageNet class labels~\footnote{\tiny \url{https://github.com/anishathalye/imagenet-simple-labels/blob/master/imagenet-simple-labels.json}} by extending with ``people'' and ``face''. 
Among these 1k labels, we mark labels as a harmful association using the taxonomy from~\cite{goyal2022fairness}, which designates classes such as gorilla as harmful label associations (see Appendix \ref{app_sec:harmful_associations} for details).
To evaluate harmful labels we use the top-5 among a models' class predictions for all our analysis and consider a prediction harmful if the majority of labels in the top-5 constitute harmful label associations. We show in Figure \ref{fig:most-common-harmful-labels} the most commonly predicted harmful labels by a CLIP ViT-L14 model. We find primates tend to be the most commonly predicted harmful label followed by ``pig'' and ``cockroach''.

\paragraph{Bias in harmful label associations.}
While we should strive to develop models that produce safe predictions (that is do not contain harmful labels) in this work we are particularly interested in the bias of harmful predictions across protected attributes such as age, gender, and apparent skin tone to understand whether harmful model predictions disproportionately affect some groups more than others. To do so, we examine the distribution of harmful labels across each group using the self-identified labels in the Casual Conversations datasets.

\section{Harmful Label Association Disparities}

We first measure the percent of harmful label associations across several models. 
We then investigate disparities in harmful label associations across age, gender, and apparent skin tone. 
Next we account for model confidence in our disparity measures to tease out whether some models predict harmful label associations with lower confidence.
Finally, we study the effect of physical adornments on harmful predictions.

\begin{figure}
    \centering
    \includegraphics[width=\linewidth]{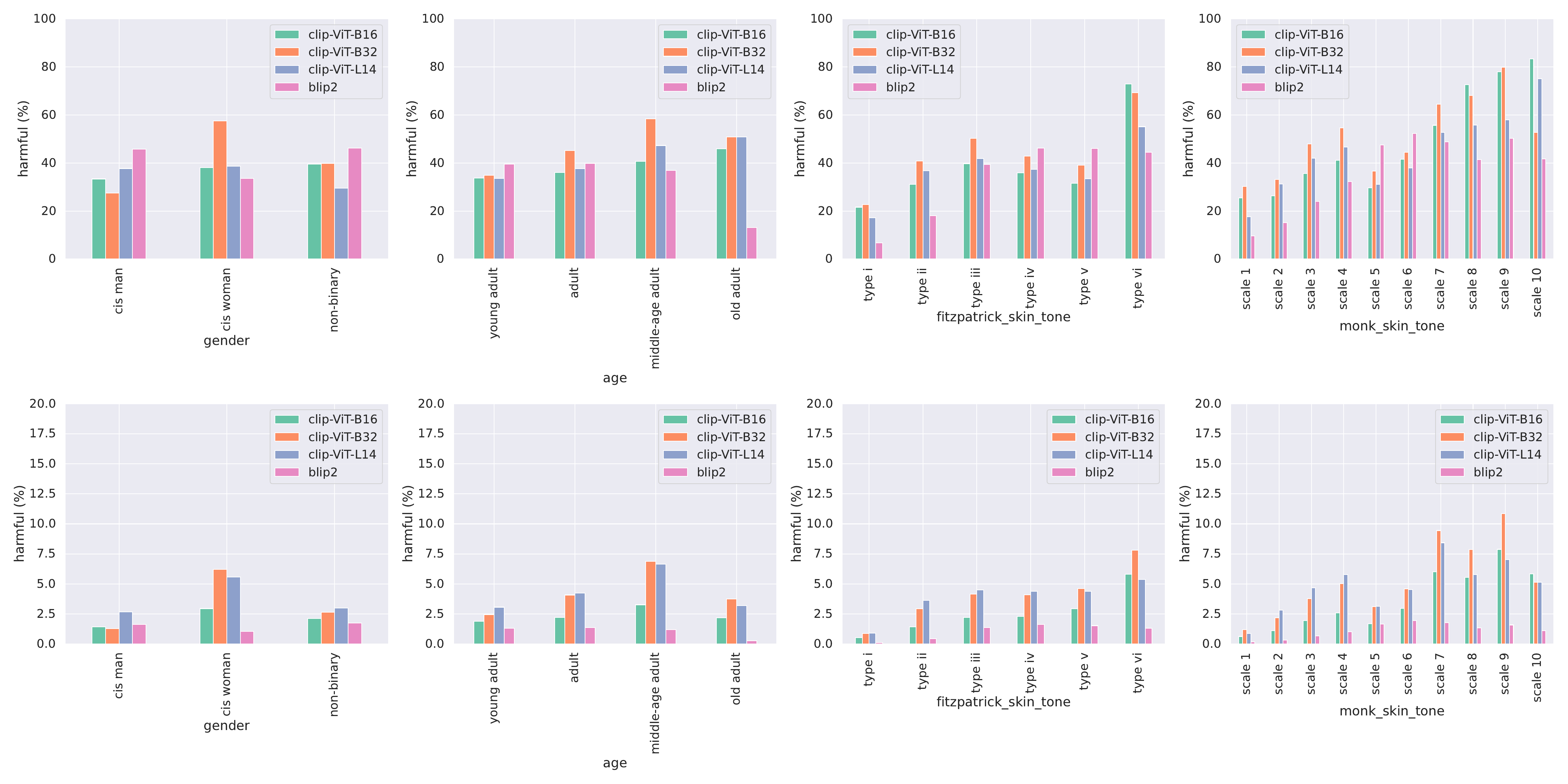}
    \caption{CCv2 breakdowns of harmful label associations. Second row shows confidence weighted scores.}
    \label{fig:harmful_assoc_ccv2}
\end{figure}

\paragraph{CLIP and BLIP2 exhibit reverse bias trends across gender and age.}
We measure the proportion of harmful label associations across gender, age, and apparent skin tones in Figure \ref{fig:harmful_assoc_ccv2}. 
For gender, we find transformer-based CLIP models exhibit a higher proportion of harmful label associations for women than for men. 
For example, CLIP ViT-B32 predicts harmful label associations for cis women at a rate of $57.5\%$ compared to only $27.5\%$ for cis men. 
The harmful association rate for non-binary individuals for CLIP models falls in-between that of cis women and cis men at $39.8\%$.
On the other hand, the BLIP-2 model predicts harmful label associations much less frequently for cis women ($33.6\%$) than cis men ($45.7\%$) and similarly non-binary individuals ($46.1\%$).
We also find reversed bias trends for CLIP versus BLIP-2 for age. While CLIP models tend predict more harmful label associations for middle-aged and older adults relative to those for adult and young adults, BLIP-2's bias is reversed. 

\paragraph{Harmful label associations are 4x more likely for darker skin tones.}
Finally, we find a stark difference in the percent of harmful label associations across apparent skin tones as show in Figure \ref{fig:harmful_assoc_ccv2} with harmful predictions occurring nearly 4x more on average for darker skin tones (type vi Fitzpatrick) compared to lighter skin tones: $72.9\%$ darker versus $21.6\%$ lighter. The disparity is consistent across all models we evaluated with BLIP-2 exhibiting a disparity in harmful label associations of $7$x across skin tones: $44.5\%$ for darker versus just $6.7\%$ for lighter. 

\paragraph{Progress on standard vision tasks does not improve disparities in harmful label associations for apparent skin tones.}
Comparing harmful association disparities for BLIP-2 and CLIP models yields an important implication for the research community. 
While BLIP-2 achieves markedly better performance across a variety of vision tasks relative to CLIP, BLIP-2's disparities in harmful label associations across skin tones are more than $2$x worse compared to those of CLIP: BLIP-2 is 7x more likely to harmfully classify individuals with apparent darker skin tones while CLIP's does so at $3$x the rate. 
All the while, on standard vision benchmarks BLIP-2 achieves markedly better results. For example, BLIP-2 achieves zero-shot retrieval of $94.9\%$ for Flick30k while CLIP only achieves $88\%$ with BLIP-2 demonstrating comparable gains on other benchmarks (COCO zero-short retrieval 85.4\% for BLIP-2 while only 77.0\% for CLIP) \citep{li2023blip2}.
This contrast between performance on standard benchmarks and disparities in harmful label associations for skin tones suggests \textit{improving performance on standard vision benchmarks does necessarily not improve disparities in harmful label associations}.

\paragraph{Some individuals are consistently harmfully classified across all videos in the dataset}
We find for nearly $4.4$\% of individuals (245 out of 5566) the same individual is harmfully associated in model predictions across all videos. To control for confounding factors such as pose, resolution or lighting, we show a randomly selected set of frames for individuals in the right panel of Figure \ref{fig:most-common-harmful-labels} as as well a random selection of frames of the same individuals across different videos who consistently harmfully associated by a CLIP ViT-L14 model in Appendix Figure \ref{app_fig:same-individual-samples}.

\subsection{Accounting for model confidence in harmful label associations}

In addition to measuring the overall proportion of harmful label associations among a model's predictions, we account for a model's confidence in such predictions.
To do so, we first compute the softmax normalized similarities between text label representations $t \in \mathbb{R}^{k \times d}$ and an image representation $z$ as \textit{normalized sim} = $\softmax(t z^\top)$. We then weigh each harmful prediction by its corresponding mean \textit{normalized sim} in the top-5 predictions. We show the resulting confidence weighted percent of harmful label associations in Figure \ref{fig:weighted_assoc_ccv2}. We surface two noteworthy trends:

\paragraph{Larger ViT models are more confident in their harmful label associations.} While model sizes across CLIP models showed no consistent trend with respect to harmful label associations when we measured the overall percent of harmful label associations, we find in contrast, CLIP models with larger encoders are much more confident in their harmful predictions relative to CLIP models with smaller encoders. This overconfidence of larger models was consistent across gender, age, and apparent skin tones.

\paragraph{BLIP in contrast to CLIP is much less confident in its harmful label associations.} Although BLIP-2 exhibited overall harmful association rates comparable or in some cases higher than those for CLIP, we find after accouting for confidence in these prediction BLIP-2 harmful prediction rates are consistently lower than those for CLIP across all model sizes. This suggests while BLIP-2 also produces harmful label associations, the model confidence is appropriately low for these harmful predictions.
As shown in Appendix Figure \ref{app_fig:harmful_assoc_ccv1}, these trends hold both for the CCv2 and CCv1 datasets.

\subsection{The effect of physical adornments on harmful label associations}

Finally, given humans may appear with various physical adornments such as glasses, make up, face masks, and hair covers, we use the self-identified labels in CCv2 to study how the presence of such physical adornments affects models' harmful predictions. We study a collection of 8 physical adornments shown in Figure \ref{fig:spurious_ccv2}. 
We find the physical adornments that cover facial features such as facial masks, eye wear, bears or mustaches considerably decrease the percent of harmful label associations. For example, the presence of a face mask decreases the percent of harmful label associations drastically across all models as shown in the top left panel of Appendix Figure \ref{fig:spurious_ccv2}. This decrease is consistent across all models suggesting models may be relying on facial features obscured by face masks such as the mouth and nose in their harmful label associations. We find a similar trend when we account for model confidence of those prediction as shown in Appendix Figure \ref{fig:weighted_assoc_ccv2}.

\section{Discussion}

We investigated disparities in models' harmful label associations across age, gender, and apparent skin tone. We find models exhibit significant bias across groups within these important axes with the most alarming trend arising for apparent skin tone: CLIP and BLIP-2 are 4-7x more likely to harmfully associate individuals with darker skin than those with lighter skin. 
We also account for model confidence, finding larger models exhibit more confidence in harmful label associations suggesting scaling models, while helpful on standard benchmarks, can exacerbate harmful label associations. Finally, we find improved performance on standard vision tasks does not necessarily correspond to improvements in harmful association disparities, suggesting addressing such disparities requires concerted research efforts with this desideratum in mind.

\bibliography{bib}
\bibliographystyle{iclr2024_conference}

\newpage

\appendix
\section{Appendix}
\subsection{Datasets\label{apx:data}}

For both datasets, we grouped the age into four buckets as ``young adult'' ($0-24$), ``adult'' ($25-40$), ``middle-age adult'' ($41-65$) and ``old adult'' ($66-100$)~\cite{hazirbas2022ccv2lit}. In CCv2, considering relatively lower number of participants, we combined ``transgender'' and ``non-binary''  as ``non-binary''. In order to run zero-shot classification, we processed the 
videos to crop faces as described in~\cite{hazirbas2022ccv1}, totalling of $45,116$ face crops/videos in CCv1 and $26,358$ face crops/videos in CCv2.

\begin{figure}
    \centering
    \includegraphics[width=\linewidth]{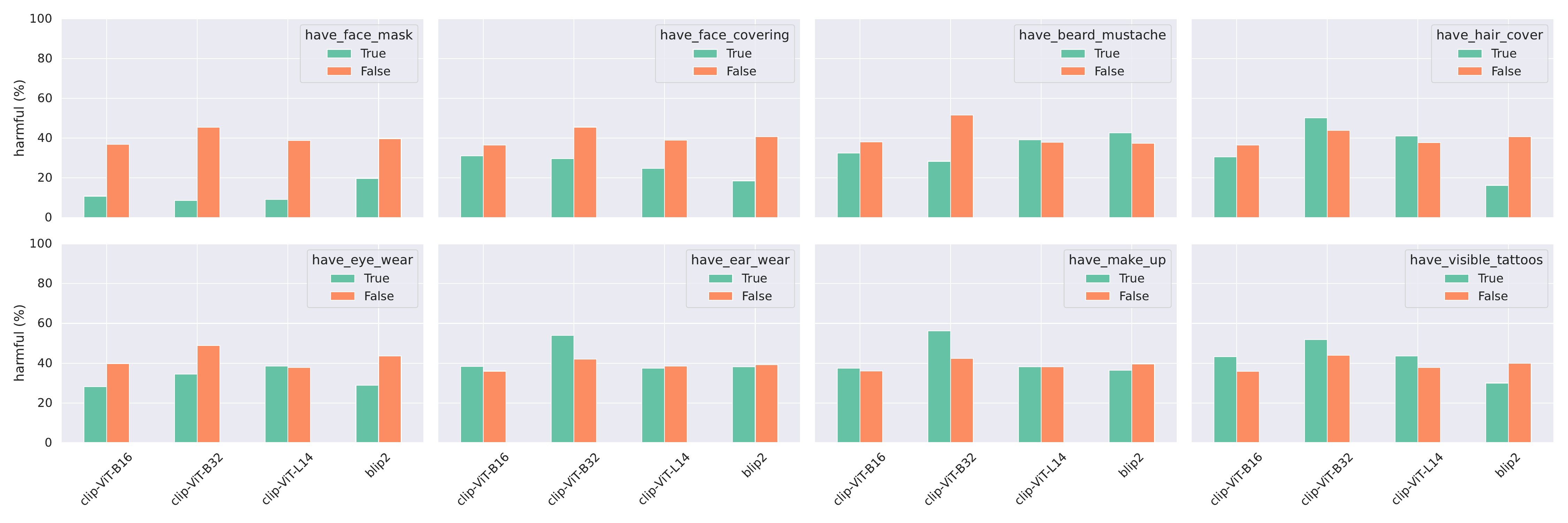}
    \caption{CCv2 physical adornments breakdowns.}
    \label{fig:spurious_ccv2}
\end{figure}

\begin{figure}
    \centering
    \includegraphics[width=\linewidth]{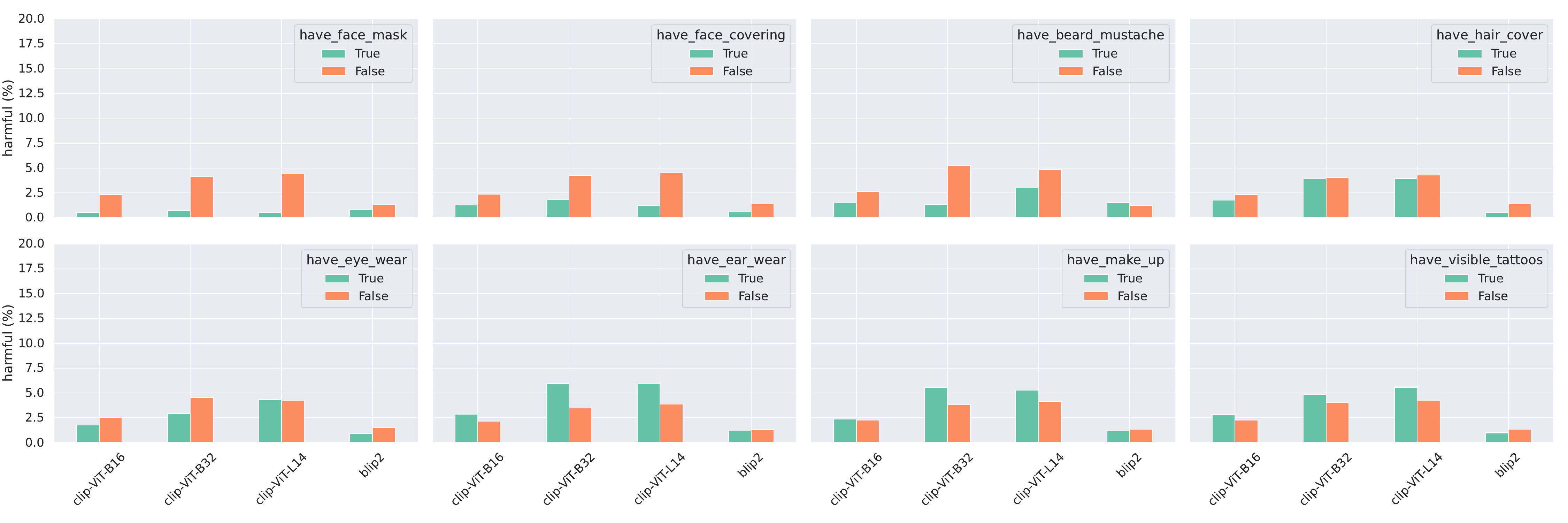}
    \caption{CCv2 physical adornments breakdowns (confidence weighted).}
    \label{fig:weighted_assoc_ccv2}
\end{figure}

\subsection{harmful label associations for CCv1}

\begin{figure}
    \centering
    \includegraphics[width=\linewidth]{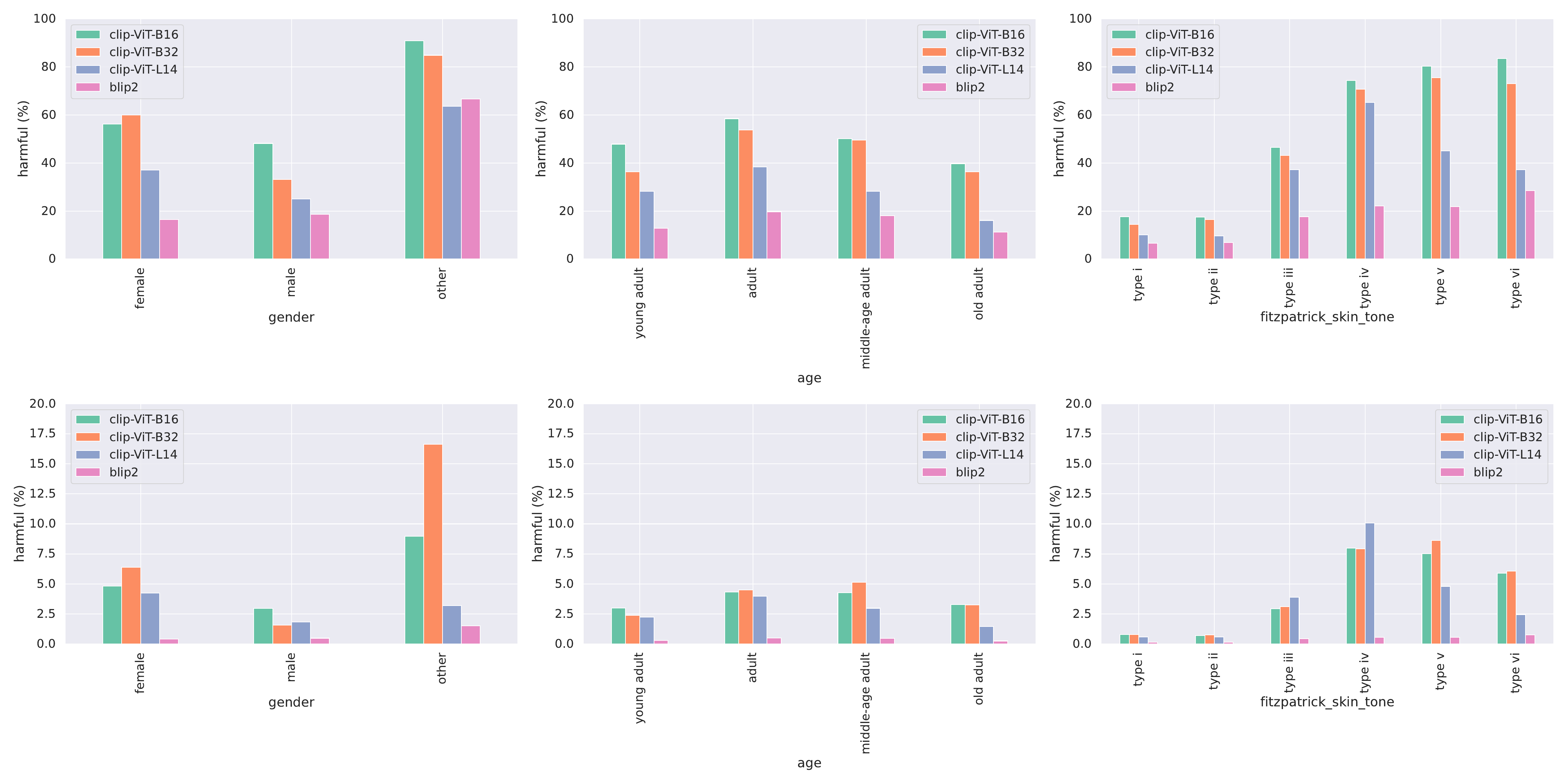}
    \caption{CCv1 breakdowns of harmful label associations. Second row shows confidence weighted scores.}
    \label{app_fig:harmful_assoc_ccv1}
\end{figure}

We also measure harmful label associations and their bias across the same set of groups for the CCv1 dataset across the models we evaluated as shown in Figure \ref{app_fig:harmful_assoc_ccv1}.

\subsection{harmful label associations}
\label{app_sec:harmful_associations}

\begin{figure}
    \centering
    \begin{tabular}{ccc}
        \includegraphics[width=0.28\linewidth]{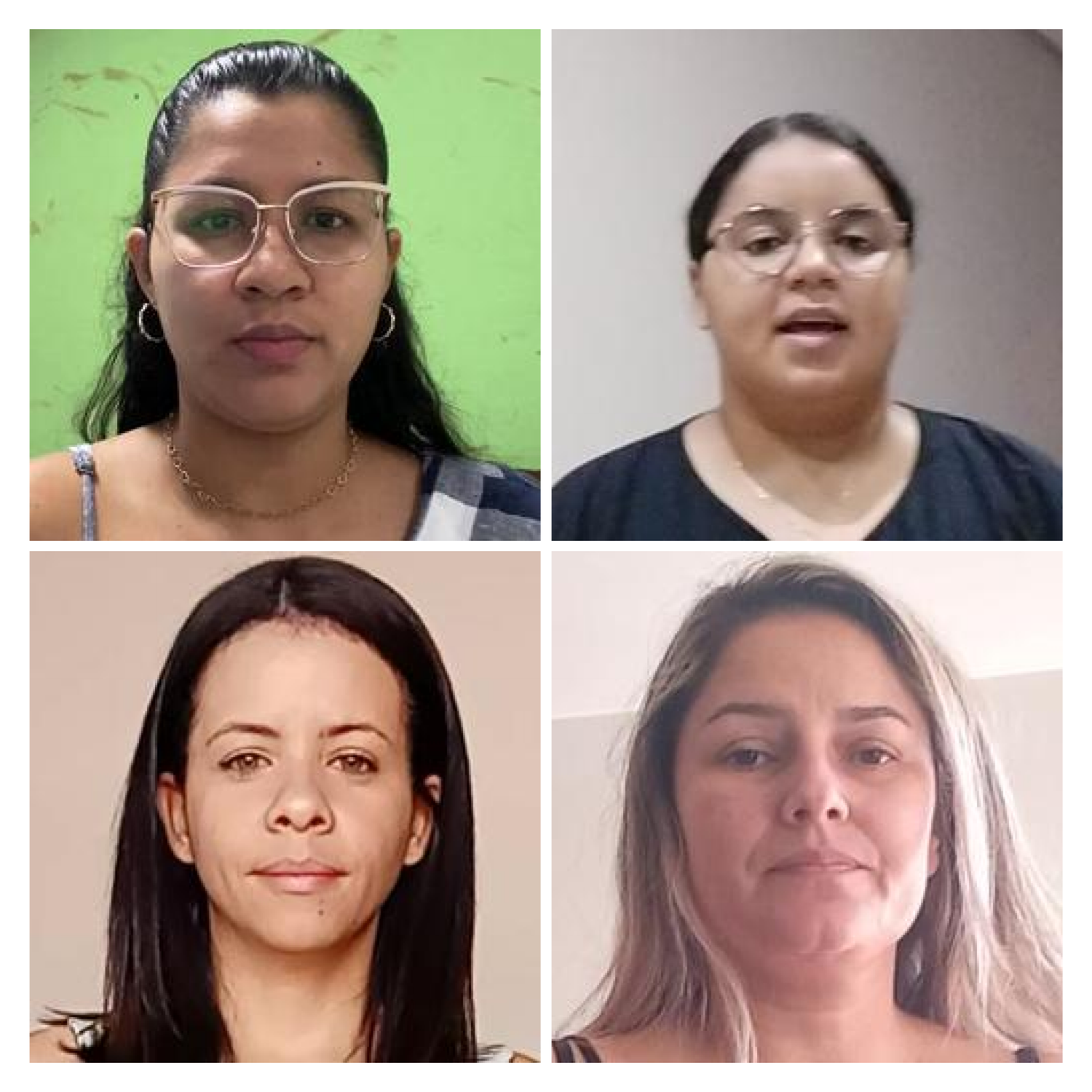} &
        \includegraphics[width=0.28\linewidth]{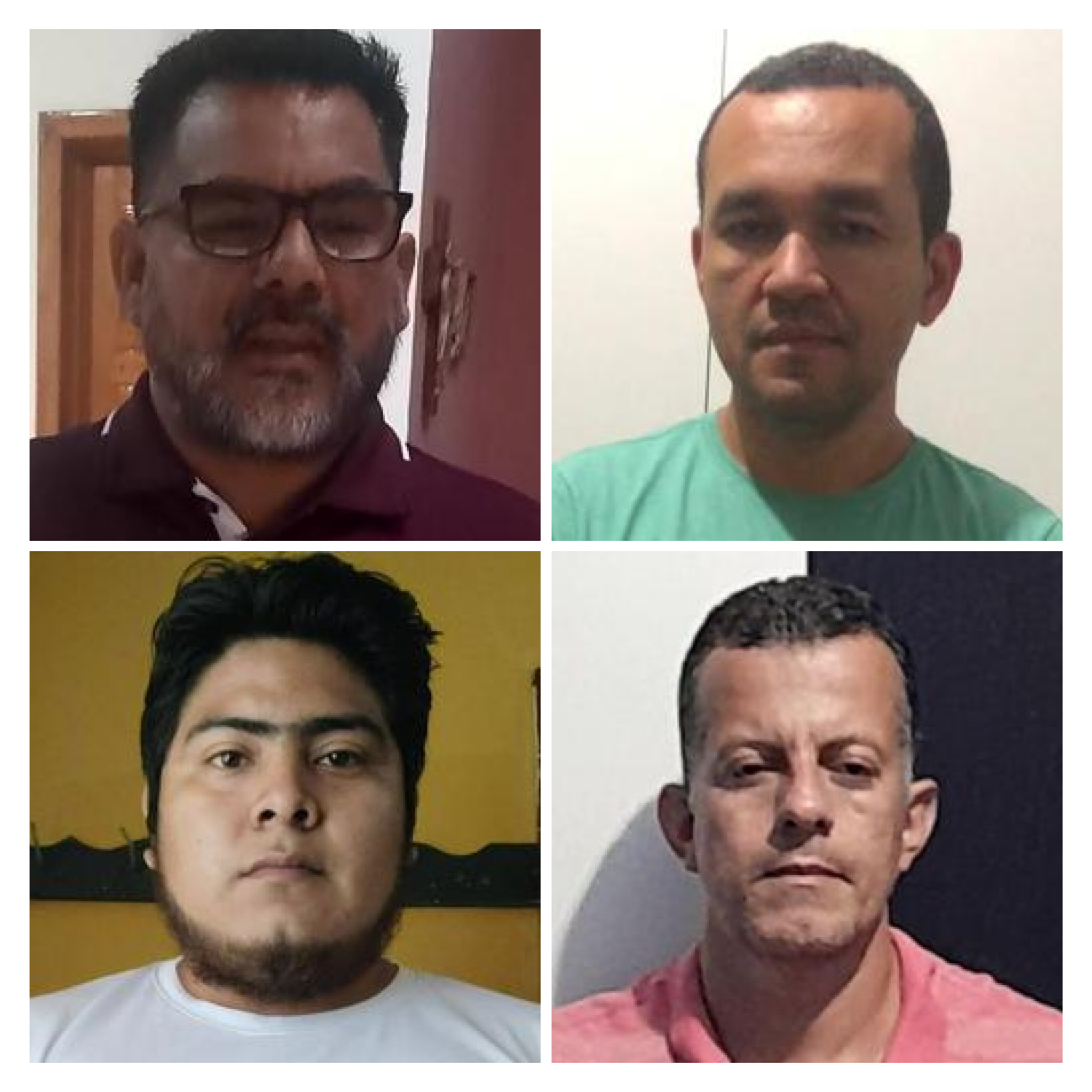} &
        \includegraphics[width=0.28\linewidth]{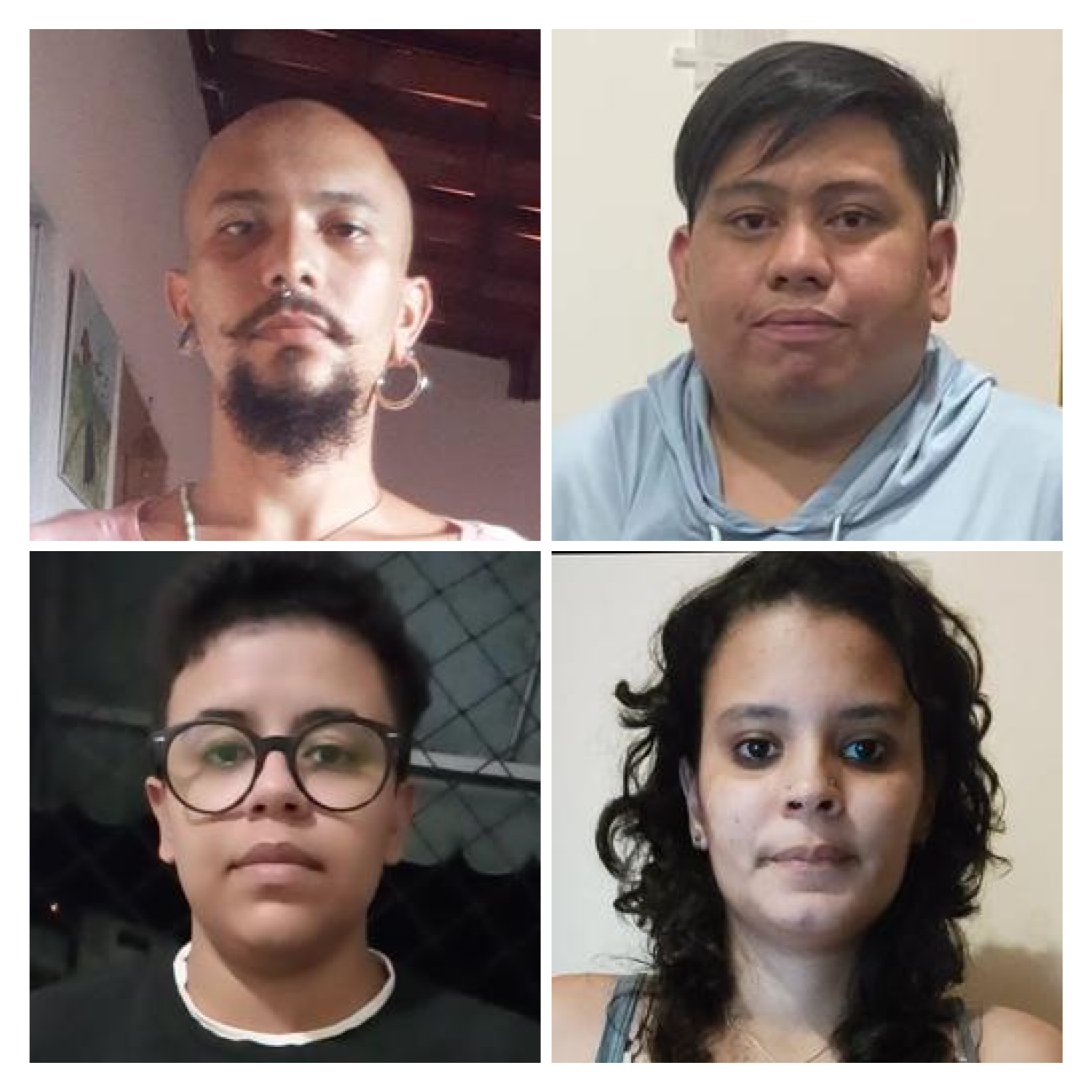}\\
         (a) Cis Woman & (b) Cis Man & (c) Transgender/Non-binary\\
    \end{tabular}
    
    \caption{Sample face crops. All videos of these participants are  harmfully associated by the CLIP ViT-L14 model.}
    \label{app_fig:individual-samples}
\end{figure}

\begin{figure}
    \centering
    \includegraphics[trim={0, 7cm, 0, 7cm},clip, width=0.93\linewidth]{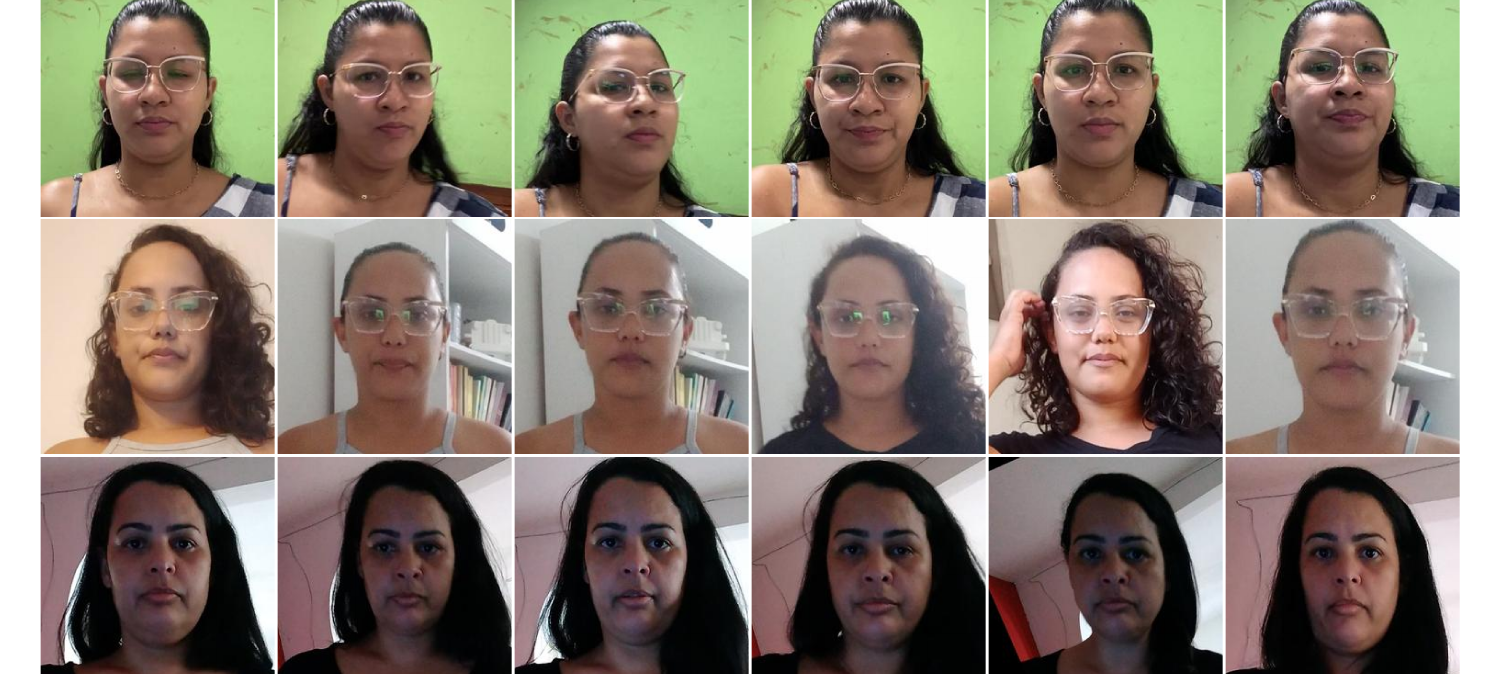}
    \caption{All of the videos of these participants classified harmful by CLIP ViT-L14.}
    \label{app_fig:same-individual-samples}
\end{figure}

\begin{figure}
    \centering
    \includegraphics[width=\linewidth]{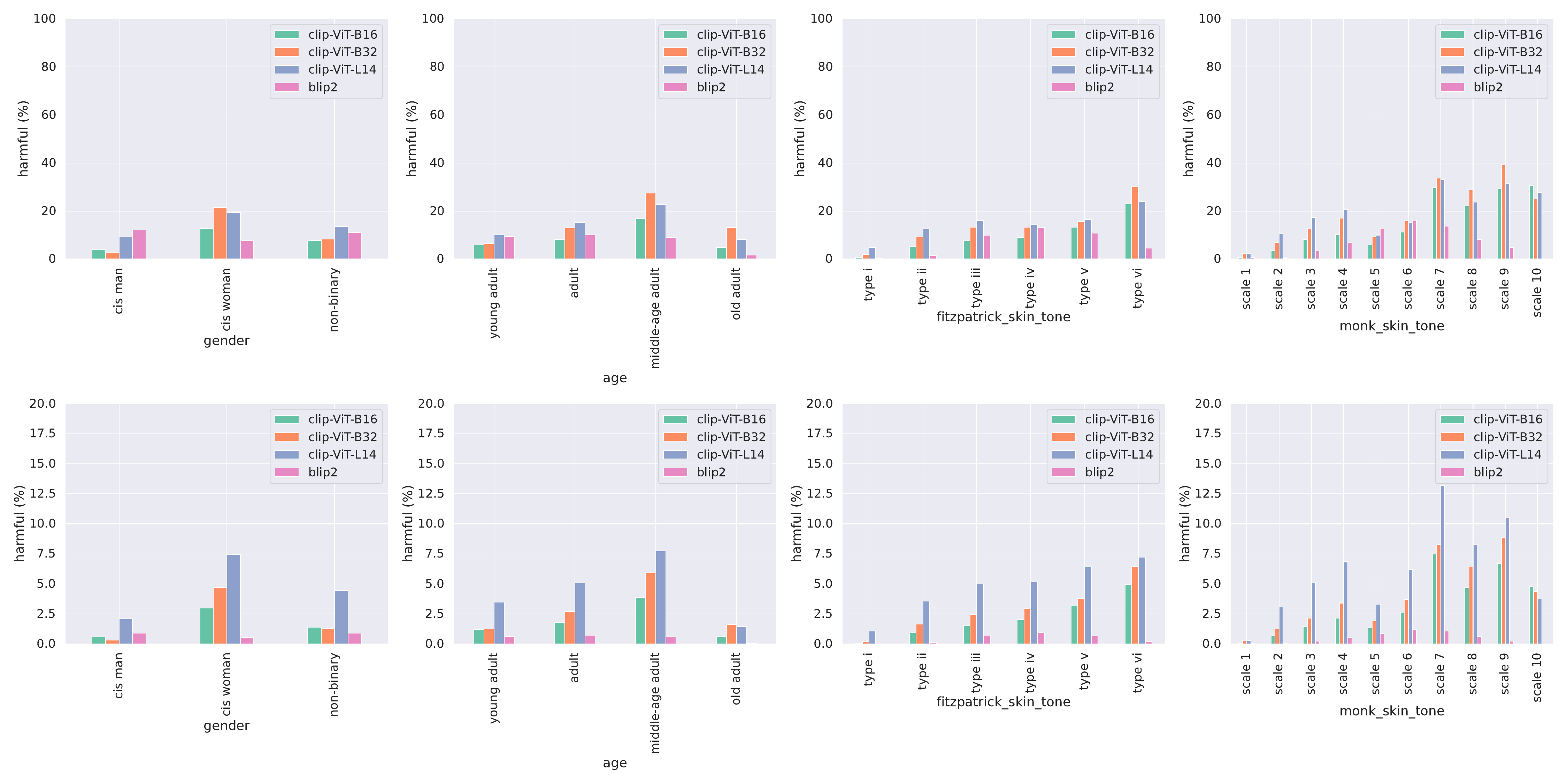}
    \caption{CCv2 breakdowns of harmful label associations on top-1. Second row shows confidence weighted scores.}
    \label{app_fig:harmful_assoc_ccv2_top1}
\end{figure}

Class label designations: 1) \textit{harmful}: orangutan, gorilla, chimpanzee, gibbon, siamang, guenon, patas monkey, baboon, macaque, langur, black-and-white colobus, proboscis monkey, marmoset, white-headed capuchin, howler monkey, titi, Geoffroy's spider monkey, common squirrel monkey, cockroach, pig; 2) \textit{possibly-human}: sunglass, sunglasses, sweatshirt, lipstick; 3) \textit{human}: people and face. Although these lists are not the final, it allows us to measure the most prominent harmful label associations in VLMs. We show sample frames for individuals across self-labeled gender categories in Figure \ref{app_fig:individual-samples} all of which are harmfully associated by a CLIP ViT-L14 model. We also show sample frames for the same individuals who are consistently harmfully associated across all videos in Figure~\ref{app_fig:same-individual-samples}. In addition, Figure~\ref{app_fig:harmful_assoc_ccv2_top1} demonstrates harmful associations on top-1 predictions.

\end{document}

%% file: math_commands.tex
\usepackage{amsmath,amsfonts,bm}

\def\eqref#1{equation~\ref{#1}}

\def\1{\bm{1}}

\DeclareMathAlphabet{\mathsfit}{\encodingdefault}{\sfdefault}{m}{sl}
\SetMathAlphabet{\mathsfit}{bold}{\encodingdefault}{\sfdefault}{bx}{n}

\newcommand{\softmax}{\mathrm{softmax}}

